\documentclass[11pt,a4paper]{article}
\usepackage{authblk}
\usepackage[hyperref]{acl2017}
\usepackage{times}
\usepackage{latexsym}
\usepackage{todonotes}
\usepackage{graphicx,multirow}
\usepackage{url}

\aclfinalcopy % Uncomment this line for the final submission
\title{Understanding Abuse:\\A Typology of Abusive Language Detection Subtasks}

\author[*]{Zeerak Waseem}
\author[$\dagger$]{Thomas Davidson}
\author[$\ddagger$]{Dana Warmsley}
\author[$\S$]{Ingmar Weber}
\affil[*]{Department of Computer Science, University of Sheffield, United Kingdom}
\affil[$\dagger$]{Department of Sociology, Cornell University, Ithaca, NY}
\affil[$\ddagger$]{Center for Applied Mathematics, Cornell University, Ithaca, NY}
\affil[$\S$]{Qatar Computing Research Institute, HBKU, Doha, Qatar}
\affil[ ]{\texttt{z.w.butt@shef.ac.uk}, \texttt{\{trd54,dw457\}@cornell.edu}, \texttt{iweber@hbku.edu.qa}}

\date{}

\begin{document}
\maketitle
%\todo{D: inserted line "hypersetup{draft}" above because we have issue with citation crossing over two pages, would not compile. I believe the problem is at the bottom of page 1.}
\begin{abstract}
  As the body of research on abusive language detection and analysis grows, there is a need for critical consideration of the relationships between different subtasks that have been grouped under this label. Based on work on hate speech, cyberbullying, and online abuse we propose a typology that captures central similarities and differences between subtasks and we discuss its implications for data annotation and feature construction. We emphasize the practical actions that can be taken by researchers to best approach their abusive language detection subtask of interest. 
  
\footnote{This paper has been accepted at the 1st Workshop on Abusive Language Online. Please be sure to cite that version.}

   %In addition, we provide a suggestion for a typology to help guide work in this area, to both indicate overlap between tasks that researchers can exploit and to avoid the conflation of multiple distinct tasks. To remedy these drawbacks, we suggest avenues and research areas for authors to consider for future work.
\end{abstract}

%\todo{T: Can we think of a better title. What about "What is abuse? A typology of abusive language detection tasks}
%\todo{Z: "Reading abuse: A typology of abusive language detection tasks". T: How about "Understanding abuse: A typology of abusive language detection tasks"... Understanding captures the ML aspect a bit better than reading}
%\todo{T: I'm commenting out todos that seem to be resolved or redundant.}
%\todo{T: We're at the point where brutal cutting down is necessary to meet the space requirement (4 pages text, 2 pages references). To this end I am going through and commenting out parts of the paper that are repetitive or no longer relevant. Since the final version, if we are accepted, will be 5 pages we can potentially include some of this material later. Apologies if I cut anything you feel strongly about, feel free to add it back in.}

\section{Introduction}

There has been a surge in interest in the detection of abusive language, hate speech, cyberbullying, and trolling in the past several years \cite{Schmidt:2017}. Social media sites have also come under increasing pressure to tackle these issues. Similarities between these subtasks have led scholars to group them together under the umbrella terms of ``abusive language'', ``harmful speech'', and ``hate speech'' \cite{Nobata:2016,faris2016understanding,Schmidt:2017} but little work has been done to examine the relationship between them.  As each of these subtasks seeks to address a specific yet partially overlapping phenomenon, we believe that there is much to gain by studying how they are related.
%In this paper, we begin by discussing the interconnections between the different subtasks pertaining to abuse and hostility online, and argue that researchers should seek to identify the points of overlap and difference between them in their work.To this effect, we propose a new typology that can aid researchers in selecting the analytical strategies most appropriate to their research goals.

%The term ``abusive language'' can encompass many things, including hate speech, cyberbullying, harassment, offensive language, and trolling. 

%The definitions used by researchers to classify various forms of abusive language are often common across distinct subtasks. 
%The above reads to me as though researchers are using the same "definitions" across the literature. 
The overlap between subtasks is illustrated by the variety of labels used in prior work. For example, in annotating for cyberbullying events, \newcite{Hee:2015b} identifies discriminative remarks (racist, sexist) as a subset of ``insults'', whereas \newcite{Nobata:2016} classifies similar remarks as ``hate speech'' or ``derogatory language''. \newcite{Waseem-Hovy:2016} only consider ``hate speech'' without regard to any potential overlap with bullying or otherwise offensive language, while \newcite{Davidson:2017} distinguish hate speech from generally offensive language. \newcite{Wulczyn:2017} annotates for personal attacks, which likely encompasses identifying  cyberbullying, hate speech, and offensive language. The lack of consensus has resulted in contradictory annotation guidelines - some messages considered as hate speech by \newcite{Waseem-Hovy:2016} are only considered derogatory and offensive by \newcite{Nobata:2016} and \newcite{Davidson:2017}. %For example, \newcite{Nobata:2016} consider ``yikes...another republiC*NT weighs in....'' to be derogatory but not hate speech and \newcite{Davidson:2017} guide their annotators to mark ``Y'all be cool out there in that cold chasing them h*es'' as offensive but not hate speech, however under the annotation guidelines proposed by \newcite{Waseem-Hovy:2016} both examples would be considered as hate speech. 
%This lack of agreement on definition seems to be partially fueled by a lack of theoretical grounding in the research fields working on the topics, in particular that of hate speech detection. The choice not to base definitions in prior research in related fields is sometimes motivated by an attempt to annotate for the community guidelines adopted by social media research \cite{Davidson:2017} or legal frameworks which define hate speech \cite{Nobata:2016}. 

To help to bring together these literatures and to avoid these contradictions, we propose a typology that synthesizes these different subtasks. We argue that the differences between subtasks within abusive language can be reduced to two primary factors:
\begin{enumerate}
 \setlength\itemsep{1em}
 \item {\textit{Is the language directed towards a specific individual or entity or is it directed towards a generalized group?}}
\item{\textit{Is the abusive content explicit or implicit?}}
\end{enumerate}

Each of the different subtasks related to abusive language occupies one or more segments of this typology. Our aim is to clarify the similarities and differences between subtasks in abusive language detection to help researchers select appropriate strategies for data annotation and modeling.

\begin{table*}[ht]
\centering
\begin{tabular}{p{\textwidth/30}|p{0.45\textwidth}|p{0.45\textwidth}}

  & \textit{Explicit}    & \textit{Implicit} \\\hline
    \multirow{4}{*}{\rotatebox[origin=c]{90}{\textit{Directed}}}    &   {\scriptsize``Go kill yourself'',  ``You're a sad little f*ck'' \cite{Hee:2015a}}, \newline {\scriptsize ``@User shut yo beaner ass up sp*c and hop your f*ggot ass back across the border little n*gga''  \cite{Davidson:2017}}, \newline {\scriptsize ``Youre one of the ugliest b*tches Ive ever fucking seen'' \cite{Kontostathis:2013}}. & {\scriptsize ``Hey Brendan, you look gorgeous today. What beauty salon did you visit?'' \cite{dinakar2012common}, \newline ``(((@User))) and what is your job?  Writing cuck articles and slurping Google balls?  \#Dumbgoogles'' \cite{Hine:2016},\newline  ``you're intelligence is so breathtaking!!!!!!'' \cite{dinakar2011modeling}}\\\hline
  \multirow{5}{*}{\rotatebox[origin=c]{90}{\textit{Generalized}}} & {\scriptsize``I am surprised they reported on this crap who cares about another dead n*gger?'', ``300 missiles are cool! Love to see um launched into Tel Aviv! Kill all the g*ys there!'' \cite{Nobata:2016}, \newline ``So an 11 year old n*gger girl killed herself over my tweets? \^ \_ \^\ thats another n*gger off the streets!!'' \cite{Kwok:2013}}. & {\scriptsize``Totally fed up with the way this country has turned into a haven for terrorists. Send them all back home.'' \cite{burnap2015cyber}, \newline ``most of them come north and are good at just mowing lawns'' \cite{dinakar2011modeling}, \newline ``Gas the skypes'' \cite{magu2017detecting}} \\
\end{tabular}
\caption{\textbf{Typology of abusive language.}}
\label{tab:top}
\end{table*}

\section{A typology of abusive language}
%Nobata's distinction between hate speech (generalized other) and derogatory speech (individual or group, but not hate speech) lends itself to our typology)
Much of the work on abusive language subtasks can be synthesized in a two-fold typology that considers whether (i) the abuse is directed at a specific target, and (ii) the degree to which it is explicit. 

Starting with the targets, abuse can either be directed towards a specific individual or entity, or it can be used towards a generalized \textit{Other}, for example people with a certain ethnicity or sexual orientation. This is an important sociological distinction as the latter references a whole category of people rather than a specific individual, group, or organization (see \citealt{brubaker2004ethnicity}, \citealt{wimmer2013ethnic}) and, as we discuss below, entails a linguistic distinction that can be productively used by researchers. To better illustrate this, the first row of Table~\ref{tab:top} shows examples from the literature of directed abuse, where someone is either mentioned by name, tagged by a username, or referenced by a pronoun.\footnote{All punctuation is as reported in original papers. We have added all the * symbols.} Cyberbullying and trolling are instances of directed  abuse,  aimed at individuals and online communities respectively. The second row shows cases with abusive expressions towards generalized groups such as racial categories and sexual orientations.   Previous work has identified instances of hate speech that are both directed and generalized \cite{burnap2015cyber,Waseem-Hovy:2016,Davidson:2017}, although  \newcite{Nobata:2016} come closest to making a distinction between directed and generalized  hate.

The other dimension is the extent to which abusive language is explicit or implicit.  This is roughly analogous to the distinction in linguistics and semiotics between \textit{denotation}, the literal meaning of a term or symbol, and \textit{connotation}, its sociocultural associations, famously articulated by \newcite{barthes1957mythologies}. 
Explicit abusive language is that which is unambiguous in its \textit{potential} to be abusive, for example language that contains racial or homophobic slurs. Previous research has indicated a great deal of variation  within such language \cite{Warner:2012,Davidson:2017}, with abusive terms being used in a colloquial manner or by people who are victims of abuse. 
%It is not necessarily explicit in its meaning but in its potential to be abusive. In other words, it may contain language that denotes abuse but the actual abusiveness depends on other factors like the social context.
Implicit abusive language is that which does not immediately imply or denote abuse. Here, the true nature is often obscured by the use of ambiguous terms, sarcasm, lack of profanity or hateful terms, and other means, generally making it more difficult to detect by both annotators and machine learning approaches \cite{dinakar2011modeling,Dadvar:2013,Justo:2014}. Social scientists and activists have recently been paying more attention to implicit, and even unconscious, instances of abuse that have been termed ``micro-aggressions'' \cite{Sue:2007}. As the examples show, such language may nonetheless have extremely abusive connotations. The first column of Table~\ref{tab:top} shows instances of explicit abuse, where it should be apparent to the reader that the content is abusive. The messages in the second column are implicit and it is harder to determine whether they are abusive without knowing the context. For example, the word ``them'' in the first two examples in the generalized and implicit cell refers to an ethnic group, and the words ``skypes" and ``Google'' are used as euphemisms for slurs about Jews and African-Americans respectively.  Abuse using sarcasm can be even more elusive for detection systems, for instance the seemingly harmless comment praising someone's intelligence was a sarcastic response to a beauty pageant contestant’s unsatisfactory answer to a question \cite{dinakar2011modeling}.

\section{Implications for future research}

In the following section we outline the implications of this typology, highlighting where the existing literatures indicate how we can understand, measure, and model each subtype of abuse. 
%We begin by discussing the implications for data annotation, then discuss the different types of features that are suited to the subtasks.
%For each cell in the typology, we outline strategies specific to the respective domains that have been used in prior research, and make suggestions for other avenues for researchers to take, which may help on the respective tasks.

\subsection{Implications for annotation}

%A significant amount of research currently fails to adequately discuss the annotation strategies used, making it difficult to assess the appropriateness of the labeling of the phenomenon it captures.  %For example, \newcite{djuric2015hate} claim to use the largest set of comments annotated for hate speech but do not indicate how these judgments were made or what they consider to be hate speech. Similarly\todo{Do Williams and Burnap explain anywhere how they went from having annotators label tweets as "hateful and/or antagonist or not" to classifying tweets as extreme, moderate or no cyberhate? I've yet to be able to find this information, perhaps I missed it, but they could be another example (so we aren't singling one paper out as not sharing).} 

In the task of annotating documents that contain bullying, it appears that there is a common understanding of what cyberbullying entails: an intentionally harmful electronic attack by an individual or group against a victim, usually repetitive in nature \cite{Dadvar:2013}. This consensus allows for a relatively consistent set of annotation guidelines across studies, most of which simply ask annotators to determine if a post contains bullying or harassment \cite{Dadvar:2014,Kontostathis:2013,Bretschneider:2014}. High inter-annotator agreement on cyberbullying tasks ($93\%$) \cite{Dadvar:2013} further indicates a general consensus around the features of cyberbullying \cite{Hee:2015b}. After bullying has been identified annotators are typically asked more detailed questions about the extremity of the bullying, the identification of phrases that indicate bullying, and the roles of users as bully/victim \cite{Dadvar:2014,Hee:2015b,Kontostathis:2013}. 

We expect that consensus may be due to the directed nature of the phenomenon. Cyberbullying involves a victim whom annotators can identify and relatively easily discern whether statements directed towards the victim should be considered abusive. In contrast, in work on annotating harassment, offensive language, and hate speech there appears to be little consensus on definitions and lower inter-annotator agreement ($\kappa \approx 0.60 - 0.80$) \cite{Ross:2016,Waseem:2016,Tulkens:2016,Bretschneider:2017} are obtained. Given that these tasks are often broadly defined and the target is often generalized, all else being equal, it is more difficult for annotators to determine whether statements should be considered abusive. %In addition, the varied understandings from academic to culturally specified understandings aid to difficulty in obtaining consistent annotations using crowd-sourcing methods.
Future work in these subtasks should aim to have annotators distinguish between targeted and generalized abuse so that each subtype can be modeled more effectively.

Annotation (via crowd-sourcing and other methods) tends to be more straightforward when explicit instances of abusive language can be identified and agreed upon \cite{Waseem-Thesis:2016}, but is considerably more difficult when implicit abuse is considered \cite{Dadvar:2013,Justo:2014,dinakar2011modeling}. The connotations of language can be difficult to classify without domain-specific knowledge. Furthermore, while some argue that detailed guidelines can help annotators to make more subtle distinctions \cite{Davidson:2017}, others find that they do not improve the reliability of non-expert classifications \cite{Ross:2016}. In such cases, expert annotators with domain specific knowledge are preferred as they tend to produce more accurate classifications \cite{Waseem:2016}. 

Ultimately, the nature of abusive language can be extremely subjective, and researchers must endeavor to take this into account when using human annotators. \newcite{Davidson:2017}, for instance, show that annotators tend to code racism as hate speech at a higher rate than sexism. As such, it is important that researchers consider the social biases that may lead people to disregard certain types of abuse. 

The type of abuse that researchers are seeking to identify should guide the annotation strategy. Where subtasks occupy multiple cells in our typology, annotators should be allowed to make nuanced distinctions that differentiate  between different types of abuse. In highlighting the major differences between different abusive language detection subtasks, our typology indicates that different annotation strategies are appropriate depending on the type of abuse. 

\subsection{Implications for modeling}

%The typology can aid in guiding research efforts in selecting features and points of analysis that have been previously observed in research. 
Existing research on abusive language online has used a diverse set of features. Moving forward, it is important that researchers clarify which features are most useful for which subtasks and which subtasks present the greatest challenges. We do not attempt to review all the features used (see \citealt{Schmidt:2017} for a detailed review) but make suggestions for which features could be most helpful for the different subtasks.  For each  aspect of the typology, we suggest features that have been shown to be successful predictors in prior work. 
% Z: ADDED
Many features occur in more than one form of abuse. As such, we do not propose that particular features are necessarily unique to each phenomenon, rather that they provide different insights and should be employed depending on what the researcher is attempting to measure.
%\T: To simply this I just removed the statement about it being overlapping. We've already said it, I think our point is clear now
%D: Due to the intersectional nature of our typology, some features will be useful in detecting more than one form of abuse. As such, we do not propose that these features are necessarily unique to each form, rather that they provide different insights and should be employed based on the insight the researcher wishes to obtain.
%Z: Due to the overlapping/interconnected nature of our typology, some features will be useful in detecting more than one form of abuse. As such, we do not propose that these features are necessarily unique to each form, rather that they provide different insights and should be employed based on the insight the researcher wishes to obtain.

\textit{Directed abuse}.
Features that help to identify the target of abuse are crucial to directed abuse detection. Mentions, proper nouns, named entities, and co-reference resolution can all be used in different contexts to identify targets. %\todo{Z: Is reference resolution the right task? I can't remember the name but it's the closest I'm getting. D: appears to be, based on a quick review of lit. T: Co-references resolution, e.g. if we have the statement "I hate X, .... they are an ****" co-reference resolution could be used to find that X and 'they' refer to the same entitity.}
% T: named entities used to identify sources and targets of hate in \cite{gitari2015lexicon}.... 
\newcite{Bretschneider:2017} use a multi-tiered system, first identifying offensive statements, then their severity, and finally the target. % of the statement. 
%As the abuse is directed, there is a chance that the target may be flagging the content for moderation. In case of using crowd-sourcing for labeling, it may prove useful to account for the directed nature in the annotation guidelines. - T: This is more relevant to the above section
Syntactical features have also proven to be successful in identifying abusive language. A number of studies on hate speech use part-of-speech sequences to model the expression of hatred \cite{Warner:2012,gitari2015lexicon,Davidson:2017}. Typed dependencies offer a more sophisticated way to capture the relationship between terms \cite{burnap2015cyber}.  Overall, there are many tools that researchers can use to model the relationship between abusive language and targets, although many of these require high-quality annotations to use as training data.

\textit{Generalized abuse}.
%Generalized abuse online tends to target people belonging to a small set of categories, primarily  racial, religious, and sexual minorities \cite{Silva:2016}.  For researchers interested in identifying abuse pertaining to a specific group, a lexical method may be an appropriate strategy.  Incorporating syntactic features also shows promise for the detection of generalized abuse.
%\todo{Check to the end of this paragraph. D: Proposed restructuring of this in notes. Z: I think yours sounds way better. T: Cleaned up a bit but looks good.}
%Additionally, researchers should consider identifying forms of abuse unique to each target group considered, as vocabularies may differ given the groups targeted. That is, the vocabularies employed for generalized abuse targeting trans-people and generalized abuse targeting Latin American people are likely to differ.
%D:
Generalized abuse online tends to target people belonging to a small set of categories, primarily  racial, religious, and sexual minorities \cite{Silva:2016}.  Researchers should consider identifying forms of abuse unique to each target group addressed, as vocabularies may depend on the groups targeted. For example, the language used to abuse trans-people and that used against Latin American people are likely to differ, both in the nouns used to denote the target group and the other terms associated with them. In some cases a lexical method may therefore be an appropriate strategy. Further research is necessary to determine if there are underlying syntactic structures associated with generalized abusive language.

\textit{Explicit abuse}
Explicit abuse, whether directed or generalized, is often indicated by specific keywords. Hence, dictionary-based approaches may be well suited to identify this type of abuse \cite{Warner:2012,Nobata:2016}, although the presence of particular words should not be the only criteria, even terms that denote abuse may be used in a variety of different ways \cite{Kwok:2013,Davidson:2017}. 
% Z added:
%\todo{Z: Something like this, Dana? T: I think the sentence below is redundant}
%Lexicons for explicit abuse would heavily emphasize the occurrence of explicit words, \newcite{Nobata:2016} generate a lexicon from terms occurring on Hatebase\footnote{www.hatebase.org} to generate counts of the number of abusive words in each document. 
Negative polarity and sentiment of the text are also likely indicators of explicit abuse that can be leveraged by researchers \cite{gitari2015lexicon}.
%\todo{D: Is there any difference in the types of key words we might look for here that distinguishes an explicit lexicon from a generalized lexicon? Perhaps explicit is inclusive of generalized but more broad - what terms might it include outside of slurs, etc? Or characteristics, if any? T: I think this is a good point. I guess the explicitness is important here... there might be a lot of words that are associated with homophobic abuse but only a small number of them are explicitly abusive (many rely on connotations)} 

%Use of certain words, dictionary-based approaches may be  appropriate. 
%Clear intention  in the speech, should be relatively easy to classify without using experts.
%Word embeddings show great promise in capturing many terms that are associated with abuse to enable extremely accurate classification \cite{djuric2015hate,badjatiya2017deep}.

\textit{Implicit abuse}.
%To analyze documents, which fit in the implicit case, documents may be hard to annotate, we propose recruiting annotators specifically trained to analyze language for the specific type of abuse considered  as proposed by \newcite{Waseem:2016}. For features, building a specific lexicon may prove difficult and impractical, as in the case of the reappropriation of the term ``skype'' in some forums \cite{magu2017detecting}, however even partial lexicons may be created and used as seeds used to inductively discover other keywords by use of a semi-supervised method proposed by \newcite{king2017computer}. Additionally, character n-grams have been shown to be apt for abusive language tasks due to their ability to capture variation of words that may be associated with abuse \cite{Nobata:2016,Waseem:2016}
%
%To combat difficulties in annotating documents that fall into the implicit category, we propose recruiting annotators specifically trained to analyze language for the specific type of abuse considered, as proposed by \newcite{Waseem:2016}. 
%T: Move this to the annotation section and integrate with discussion.
%
Building a specific lexicon may prove impractical, as in the case of the appropriation of the term ``skype'' in some forums \cite{magu2017detecting}. Still, even partial lexicons may be used as seeds to inductively discover other keywords by use of a semi-supervised method proposed by \newcite{king2017computer}. Additionally, character n-grams have been shown to be apt for abusive language tasks due to their ability to capture variation of words associated with abuse \cite{Nobata:2016,Waseem:2016}. Word embeddings are also promising ways to capture terms associated with abuse \cite{djuric2015hate,badjatiya2017deep}, although they may still be insufficient for cases like 4Chan's connotation of ``skype'' where a word has a dominant meaning and a more subversive one. Furthermore, as some of the above examples show, implicit abuse often takes on complex linguistic forms like sarcasm, metonymy, and humor. Without high quality labeled data to learn these representations, it may be difficult for researchers to come up with models of syntactic structure that can help to identify implicit abuse. To overcome these limitations researchers may find it prudent to incorporate features beyond just textual analysis, including the characteristics of the individuals involved  \cite{Dadvar:2013} and other extra-textual features. %, as is often done in cyberbullying and trolling literature (/cite{Dadvar:2013}

%If euphemisms/neologisms are used then can generate  a specific lexicon, but may be difficult and impractical depending on the euphemism (4Chan case).  A single seed keyword can be used to inductively discover other keywords using a semi-supervised method proposed by \cite{king2017computer}. Character n-gram models may also be suitable as they can capture variations of common words that may be associated with abuse \cite{Nobata:2016,Waseem:2016}. 

\section{Discussion}

% Break down the various approaches and to their implications and consequences here.
%%T: Moving this to the annotation section
This typology has a number of implications for future work in the area. 

First, we want to encourage researchers working on these subtasks to learn from advances in other areas. Researchers working on purportedly distinct subtasks are often working on the same problems in parallel. For example, the field of hate speech detection can be strengthened by interactions with work on cyberbullying, and vice versa, since a large part of both subtasks consists of identifying targeted abuse.

Second, we aim to highlight the important distinctions within subtasks that have hitherto been ignored. For example, in much hate speech research, diverse types of abuse have been lumped together under a single label, forcing models to account for a large amount of within-class variation. We suggest that fine-grained distinctions along the axes allows for more focused systems that may be more effective at identifying particular types of abuse. %, although we expect that learning one subtask may enable models to perform better on other subtasks. 

Third, we call for closer consideration of how annotation guidelines are related to the phenomenon of interest. The type of annotation and even the choice of annotators should be motivated by the nature of the abuse.  Further, we welcome discussion of annotation guidelines and the annotation process in published work. Many existing studies only tangentially mention these, sometimes never explaining how the data were annotated. 

Fourth,  we  encourage researchers to consider which features are most appropriate for each subtask. Prior work has found a diverse array of features to be useful in understanding and identifying abuse, but we argue that different feature sets will be relevant to different subtasks. Future work should aim to build a more robust understanding of when to use which types of features.

Fifth, it is important to emphasize that not all abuse is equal, both in terms of its effects and its detection. We expect that social media and website operators will be more interested in identifying and dealing with explicit abuse, while activists, campaigners, and journalists may have more incentive to also identify implicit abuse. Targeted abuse such as cyberbullying may be more likely to be reported by victims and thus acted upon than generalized abuse. We also expect that implicit abuse will be more difficult to detect and model, although methodological advances may make such tasks more feasible.

%Further, the analysis and annotation of abusive language cannot be divorced from the form the abusive language takes. Systems that aim to detect one form of abusive language use may differ significantly from systems that seek to address another form. For instance, systems which are developed to detect and analyze generalized and implicit abusive language may work with different assumptions and key methods as compared to systems that aim to detect and analyze generalized explicit content.

\section{Conclusion}
We have presented a typology that synthesizes the different subtasks in abusive language detection. Our aim is to bring together findings in these different areas and to clarify the key aspects of abusive language detection. There are important analytical distinctions that have been largely overlooked in prior work and through acknowledging these and their implications we hope to improve abuse detection systems and our understanding of abusive language. 

Rather than attempting to resolve the ``definitional quagmire'' \cite{faris2016understanding} involved in neatly bounding and defining each subtask we encourage researchers to think carefully about the phenomena they want to measure and the appropriate research design.  We intend for our typology to be used both at the stage of data collection and annotation and the stage of feature creation and modeling. We hope that future work will be more transparent in discussing the annotation and modeling strategies used, and will closely examine the similarities and differences between these subtasks through empirical analyses.

\bibliography{acl2017}

\end{document}